\documentclass{article}

\usepackage{natbib}
\usepackage{microtype}
\usepackage{graphicx}
\usepackage{subfigure}
\usepackage{booktabs} 
\usepackage{printlen}
\usepackage{hyperref}

\usepackage[accepted]{icml2019}
\usepackage{url}            
\usepackage{booktabs}       
\usepackage{amsfonts}       
\usepackage{nicefrac}       
\usepackage{microtype}      
\usepackage{comment}
\usepackage{float}
\usepackage{longtable}
\usepackage{todonotes}
\usepackage{caption}
\usepackage{adjustbox}

\usepackage{amsthm}
\usepackage{amsmath}
\usepackage{amsfonts}
\usepackage{amssymb}
\usepackage{comment}
\usepackage{csquotes}
\usepackage[framemethod=tikz]{mdframed}
\usepackage{mathtools}
\usepackage{multirow}

\usepackage{appendix}
\usepackage{chngcntr}
\usepackage{etoolbox}

\usepackage{lettrine}

\usepackage{tabularx}

\usepackage{listings}
\usepackage{color}
\usepackage{enumitem}
\usepackage{makecell}

\newcommand{\bb}[1]{\mathbf{#1}}

\newcommand{\bx}{\bb{x}}

\newcommand{\bz}{\bb{z}}

\newcommand{\bT}{\boldsymbol{\theta}}

\newcommand{\kl}[2]{D_{\mathrm{KL}}\left(#1 ~ \| ~ #2\right)}

\DeclareMathOperator{\EX}{\mathbb{E}}
\newcommand{\loss}{\mathcal{L}}
\newcommand{\Mod}[1]{\ (\mathrm{mod}\ #1)}
\DeclarePairedDelimiter\floor{\lfloor}{\rfloor}

\newcommand\numberthis{\addtocounter{equation}{1}\tag{\theequation}}

\icmltitlerunning{Bit-Swap}

\begin{document}

\twocolumn[



\icmltitle{Bit-Swap: Recursive Bits-Back Coding for Lossless Compression with Hierarchical Latent Variables}



\icmlsetsymbol{equal}{*}

\begin{icmlauthorlist}
\icmlauthor{Friso H. Kingma}{uc}
\icmlauthor{Pieter Abbeel}{uc}
\icmlauthor{Jonathan Ho}{uc}
\end{icmlauthorlist}

\icmlaffiliation{uc}{University of California, Berkeley, California, USA}

\icmlcorrespondingauthor{Friso H. Kingma}{fhkingma@gmail.com}
\icmlcorrespondingauthor{Jonathan Ho}{jonathanho@berkeley.edu}

\icmlkeywords{Machine Learning, ICML}

\vskip 0.3in
]



\printAffiliationsAndNotice{}  

\begin{abstract}
The bits-back argument suggests that latent variable models can be turned into lossless compression schemes. Translating the bits-back argument into efficient and practical lossless compression schemes for general latent variable models, however, is still an open problem. Bits-Back with Asymmetric Numeral Systems (BB-ANS), recently proposed by \citet{townsend2018practical}, makes bits-back coding practically feasible for latent variable models with one latent layer, but it is inefficient for hierarchical latent variable models. In this paper we propose Bit-Swap, a new compression scheme that generalizes BB-ANS and achieves strictly better compression rates for hierarchical latent variable models with Markov chain structure. Through experiments we verify that Bit-Swap results in lossless compression rates that are empirically superior to existing techniques. Our implementation is available at \url{https://github.com/fhkingma/bitswap}. 
\end{abstract}

\section{Introduction} \label{introduction}

Likelihood-based generative models---models of joint probability distributions trained by maximum likelihood---have recently achieved large advances in density estimation performance on complex, high-dimensional data. Variational autoencoders \citep{kingma2013auto,rezende2014stochastic,kingma2016improving}, PixelRNN and PixelCNN and their variants \citep{oord2016pixel, oord2016conditional,salimans2017pixelcnn++,parmar2018image,chen2017pixelsnail}, and flow-based models like RealNVP \citep{dinh2014nice, dinh2016density, kingma2018glow} can successfully model high dimensional image, video, speech, and text data \citep{karras2017progressive, kalchbrenner2016video, oord2016wavenet, kalchbrenner2016language, kalchbrenner2018efficient, vaswani2017attention}.

The excellent density estimation performance of likelihood-based models suggests another application: lossless compression \cite{gregor2016towards}. Any distribution can in theory be converted into a lossless code, in which each datapoint is encoded into a number of bits equal to its negative log probability assigned by the model. Since the best expected codelength is achieved when the model matches the true data distribution, designing a good lossless compression algorithm is a matter of jointly solving two problems:
\begin{enumerate}
    \item Approximating the true data distribution $p_{\text{data}}(\bx)$ as well as possible with a model $p_{\bT}(\bx)$
    \item Developing a practical compression algorithm, called an \emph{entropy coding} scheme, that is compatible with this model and results in codelengths equal to~$-\log p_{\bT}(\bx)$.
\end{enumerate}


\begin{figure}[t!]
\begin{center}
\includegraphics[width=0.92\columnwidth]{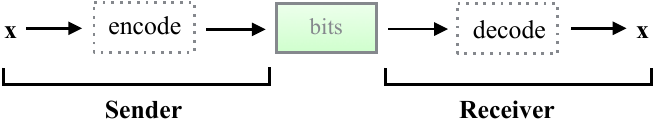}
\caption{Schematic overview of lossless compression. The sender encodes data $\bx$ to a code with the least amount of bits possible without losing information. The receiver decodes the code and must be able to exactly reconstruct $\bx$.}
\label{fig:losslesscompression}
\end{center}
\end{figure}

\begin{table}[!hb]
\caption{Lossless compression rates (in bits per dimension) on unscaled and cropped ImageNet of Bit-Swap against other compression schemes. See Section~\ref{sec:bitswap} for an explanation of Bit-Swap and Section~\ref{sec:experiments} for detailed results.}
\centering
\begin{adjustbox}{width=0.912\columnwidth}
\begin{tabular}{@{}lll@{}}
\toprule
\multirow{8}{*}{\includegraphics[width=0.50\linewidth]{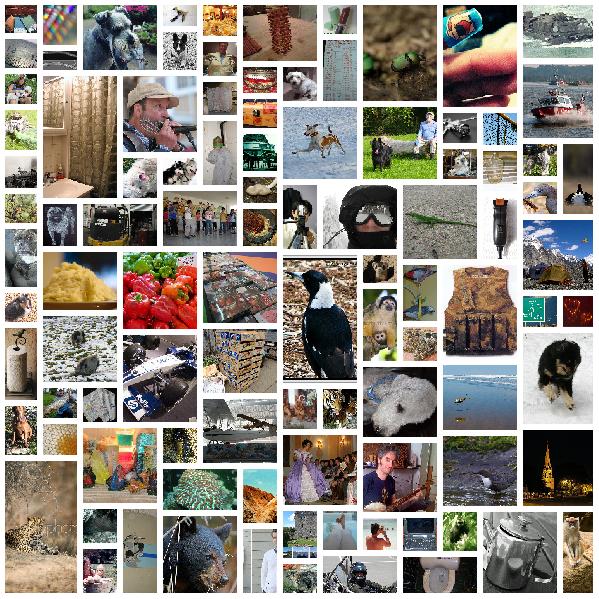}} & \textbf{Compression Scheme}           & \textbf{Rate} \\ \cmidrule(l){2-3} 
                           & Uncompressed              & 8.00             \\ \cmidrule(l){2-3} 
                           & GNU Gzip                  & 5.96             \\
                           & bzip2                     & 5.07             \\
                           & LZMA                      & 5.09             \\
                           & PNG                       & 4.71             \\
                           & WebP                      & 3.66             \\
                           & BB-ANS                    & 3.62             \\
                           & \textbf{Bit-Swap (ours)} & \textbf{3.51}    \\ \bottomrule
\end{tabular}
\end{adjustbox}
\end{table}

Unfortunately, it is generally difficult to jointly design a likelihood-based model and an entropy coding scheme that together achieve a good compression rate while remaining computationally efficient enough for practical use.
Any model with tractable density evaluation can be converted into a code using Huffman coding, but building a Huffman tree requires resources that scale exponentially with the dimension of the data.
The situation is more tractable, but still practically too inefficient, when autoregressive models are paired with arithmetic coding or asymmetric numeral systems (explained in Section \ref{sec:entropy}). The compression rate will be excellent due to to the effectiveness of autoregressive models in density estimation, but the resulting decompression process, which is essentially identical to the sample generation process, will be extremely slow.

Fortunately, fast compression and decompression can be achieved by pairing variational autoencoders with a recently proposed practically efficient coding method called Bits-Back with Asymmetric Numeral Systems (BB-ANS) \citep{townsend2018practical}. However, the practical efficiency of BB-ANS rests on two requirements: 
\begin{enumerate}
    \item All involved inference and recognition networks are fully factorized probability distributions
    \item There are few latent layers in the variational autoencoder.
\end{enumerate}
The first requirement ensures that encoding and decoding is always fast and parallelizable. The second, as we will discuss later, ensures that BB-ANS achieves an efficient bitrate: it turns out that BB-ANS incurs an overhead that grows with the number of latent variables. But these requirements restrict the capacity of the variational autoencoder and pose difficulties for density estimation performance, and hence the resulting compression rate suffers.

To work toward designing a computationally efficient compression algorithm with a good compression rate, we propose Bit-Swap, which improves BB-ANS's performance on hierarchical latent variable models with Markov chain structure. Compared to latent variables models with only one latent layer, these hierarchical latent variable models allow us to achieve better density estimation performance on complex high-dimensional distributions. Meanwhile, Bit-Swap, as we show theoretically and empirically, yields a better compression rate compared to BB-ANS on these models due to reduced overhead. 

\section{Background}
\label{background}
First, we will set the stage by introducing the lossless compression problem. Let $p_\mathrm{data}$ be a distribution over discrete data $\bx = (x_1, \dots, x_D)$. Each component $x_1, \dots, x_D$ of $\bx$ is called a \textit{symbol}. Suppose that a \textit{sender} would like to communicate a sample $\bx$ to a  \textit{receiver} through a \textit{code}, or equivalently a \textit{message}. The goal of lossless compression is to send this message using the minimum amount of bits on average over $p_\mathrm{data}$, while ensuring that $\bx$ is always fully recoverable by the receiver. See Figure~\ref{fig:losslesscompression} for an illustration.



\textit{Entropy coding} schemes use a probabilistic model $p_{\bT}(\bx)$ to define a code with codelengths $-\log p_{\bT}(\bx)$. If $-\log p_{\bT}(\bx)$ matches $-\log p_\mathrm{data}(\bx)$ well, then resulting average codelength $\mathbb{E}[-\log p_{\bT}(\bx)]$ will be close to the entropy of the data $H(\bx)$, which is the average codelength of an optimal compression scheme.

\subsection{Asymmetric Numeral Systems} \label{sec:entropy}
 We will employ a particular entropy coding scheme called \textit{Asymmetric Numeral Systems} (ANS) \cite{duda2009asymmetric}. Given a univariate probability distribution $p(x)$, ANS encodes a symbol $x$ into a sequence of bits, or \textit{bitstream}, of length approximately $-\log p(x)$ bits. 
ANS can also code a vector of symbols $\bx$ using a fully factorized probability distribution $p(\bx) = \prod_i p(x_i)$, resulting in $-\log p(\bx)$ bits. (It also works with autoregressive $p(\bx)$, but throughout this work we will only use fully factorized models for parallelizability purposes.)



\begin{figure}[t!]
\begin{center}
\includegraphics[width=0.87\columnwidth]{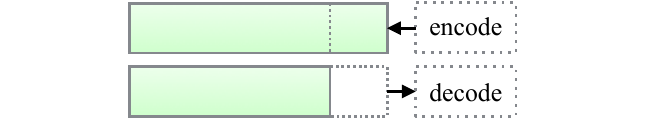}
\caption{Asymmetric Numeral Systems (ANS) operates on a bitstream in a stack-like manner. Symbols are decoded in opposite order as they were encoded.}
\label{fig:ans}
\end{center}
\end{figure}

ANS has an important property: if a sequence of symbols is encoded, then they must be decoded in the opposite order. In other words, the state of the ANS algorithm is a bitstream with a \textit{stack} structure. Every time a symbol is encoded, bits are pushed on to the right of the stack; every time a symbol is decoded, bits are popped from the right of the stack. See Figure~\ref{fig:ans} for an illustration.
This property will become important when ANS is used in BB-ANS and Bit-Swap for coding with latent variable models. 

\subsection{Latent Variable Models} \label{sec:latent}
The codelength of an entropy coding technique depends on how well its underlying model $p_{\bT}(\bx)$ approximates the true data distribution $p_{\text{data}}(\bx)$. In this paper, we focus on latent variable models, which approximate $p_{\text{data}}(\bx)$ with a marginal distribution $p_{\bT}(\bx)$ defined by 
\begin{equation}
    p_{\bT}(\bx) = \int p_{\bT}(\bx, \bz) \text{d}\bz = \int p_{\bT}(\bx|\bz)p(\bz) \text{d}\bz 
\end{equation}
where $\bz$ is an unobserved latent variable. For continuous $\bz$, $p_{\bT}(\bx)$ can be seen as an infinite mixture, which makes such an implicit distribution over $\bx$ potentially highly flexible. 

Since exactly evaluating and optimizing the marginal likelihood $p_{\bT}(\bx)$ is intractable, variational autoencoders introduce an \textit{inference model} $q_{\bT}(\bz|\bx)$, which approximates the model posterior $p_{\bT}(\bz|\bx)$. For any choice of $q_{\bT}(\bz|\bx)$, we can rewrite the marginal likelihood $p_{\bT}(\bx)$ as follows:
\begin{align}
    \log p_{\bT}(\bx) = \underbrace{\EX_{q_{\bT}(\bz | \bx)}{\log \frac{p_{\bT}(\bx, \bz)}{q_{\bT}(\bz | \bx)}}}_{=\loss(\bT) \ (\text{ELBO})} + \underbrace{ \EX_{q_{\bT}(\bz | \bx)}{\log \frac{q_{\bT}(\bx, \bz)}{p_{\bT}(\bz | \bx)}}}_{=\kl{q_{\bT}(\bz | \bx)}{p_{\bT}(\bz | \bx)}}
\end{align}
As $\kl{q_{\bT}(\bz | \bx)}{p_{\bT}(\bz | \bx)} \geq 0$, the inference model and generative model can be found by jointly optimizing the Evidence Lower BOund (ELBO), a lower bound on $\log p_{\bT}(\bx)$:
\begin{equation}
    \loss(\bT) = \EX_{q_{\bT}(\bz | \bx)}\lbrack{\log p_{\bT}(\bx, \bz)-\log q_{\bT}(\bz | \bx)\rbrack}
\end{equation}
For continuous $\bz$ and a differentiable inference model and generative model, the ELBO can be optimized using the reparameterization trick \cite{kingma2013auto}.


\subsection{Bits-Back Coding with ANS} \label{sec:bitsbackans}
It is not straightforward to use a latent variable model for compression, but it is possible with the help of the inference network $q_{\bT}(\bz | \bx)$. Assume that both the sender and receiver have access to $p_{\bT}(\bx|\bz)$, $p(\bz)$, $q_{\bT}(\bz|\bx)$ and an entropy coding scheme. Let $\bx$ be the datapoint the sender wishes to communicate. The sender can send a latent sample $\bz \sim q_{\bT}(\bz|\bx)$ by coding using the prior $p(\bz)$, along with $\bx$, coded with $p_{\bT}(\bx|\bz)$. This scheme is clearly valid and allows the receiver to recover the $\bx$, but results in an inefficient total codelength of $\EX{\lbrack -\log p_{\bT}(\bx|\bz) -\log p(\bz) \rbrack}$. \citet{wallace1990classification} and \citet{hinton1993keeping} show in a thought experiment, called the bits-back argument, it is possible to instead transmit $-\log q_\theta(\bz|\bx)$ fewer bits in a certain sense, thereby yielding a better net codelength equal to the negative ELBO $-\loss(\bT)$ of the latent variable model.

BB-ANS \cite{townsend2018practical}, illustrated in Figure~\ref{bits-back}, makes the bits-back argument concrete. BB-ANS operates by starting with ANS initialized with a bitstream of $N_\mathrm{init}$ random bits. Then, to encode $\bx$, BB-ANS performs the following steps:

\begin{enumerate} \label{bits-back}
    \item Decode $\bz$ from bitstream using $q_{\bT}(\bz|\bx)$, subtracting $-\log q_{\bT}(\bz|\bx)$ bits from the bitstream,
    \item Encode $\bx$ to bitstream using $p_{\bT}(\bx|\bz)$, adding  $-\log p_{\bT}(\bx|\bz)$ bits to the bitstream,
    \item Encode $\bz$ to bitstream using $p(\bz)$, adding $-\log p(\bz)$ bits to the bistream.
\end{enumerate}

The resulting bitstream, which has a length of $N_\mathrm{total} \coloneqq N_\mathrm{init} + \log q_{\bT}(\bz|\bx) -\log p_{\bT}(\bx|\bz) - \log p(\bz)$ bits, is then sent to the receiver.

\begin{figure}[t!]
\begin{center}
\includegraphics[width=0.80\columnwidth]{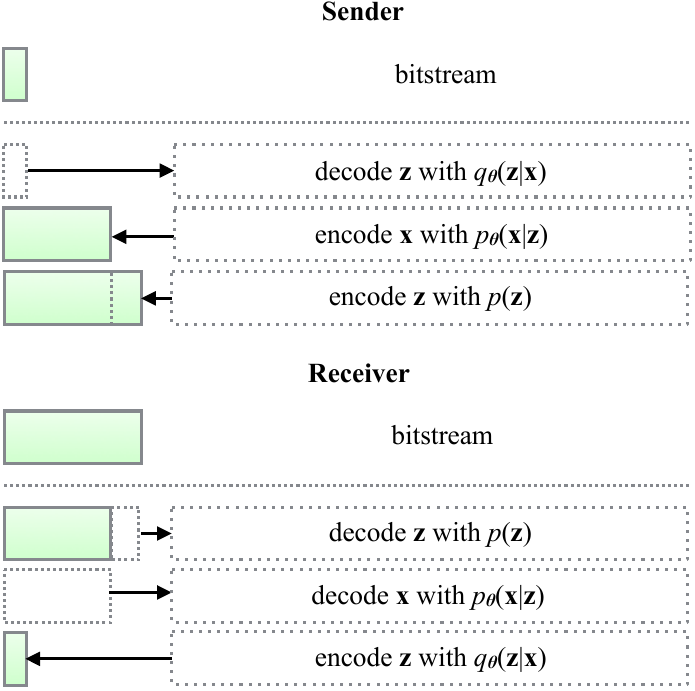}
\caption{Bits-Back with Asymmetric Numeral Systems (BB-ANS).}
\label{fig:bitsback}
\end{center}
\end{figure}

The receiver decodes the data by initializing ANS to the received bitstream, then proceeds in reverse order, with the encode and decode operations swapped: the receiver decodes $\bz$ using $p(\bz)$, decodes $\bx$ using $p_{\bT}(\bx|\bz)$, then encodes $\bz$ using $q_{\bT}(\bz|\bx)$. The final step of encoding $\bz$ will recover the $N_\mathrm{init}$ bits that the encoder used to initialize ANS. Thus, the sender will have successfully transmitted $\bx$ to the receiver, along with the initial $N_\mathrm{init}$ bits---and it will have taken $N_\mathrm{total}$ bits to do so.

To summarize, it takes $N_\mathrm{total}$ bits to transmit $\bx$ plus $N_\mathrm{init}$ bits. In this sense, the \emph{net} number of bits sent regarding $\bx$ only, ignoring the initial $N_\mathrm{init}$ bits, is
\[ N_\mathrm{total} - N_\mathrm{init} = \log q_{\bT}(\bz|\bx) -\log p_{\bT}(\bx|\bz) - \log p(\bz) \]
which is on average equal to $-\loss(\bT)$, the negative ELBO.

\section{Initial Bits in Bits-Back Coding}
\label{sec:bitsbackoverhead}

We now turn to the core issue that our work addresses: the amount of initial bits $N_\mathrm{init}$ required for BB-ANS to function properly.

It is crucial for there to be enough initial bits in the ANS state for the sender to decode $\bz$ from the initial bitstream. That is, we must have
\begin{align}N_\mathrm{init} \geq - \log q_{\bT}(\bz|\bx) \end{align}
in order to guarantee that the receiver can recover the initial $N_\mathrm{init}$ bits. If not, then to sample $\bz$, the sender must draw bits from an auxiliary random source, and those bits will certainly not be recoverable by the receiver. And, if those bits are not recoverable, then the sender will have spent $N_\mathrm{total}$ bits to transmit $\bx$ \emph{only}, without $N_\mathrm{init}$ bits in addition. So, we must commit to sending at least $N_\mathrm{init} \geq - \log q_{\bT}(\bz|\bx)$ initial bits to guarantee a short net codelength for $\bx$.

Unfortunately, the initial number of bits required can be significant. As an example, if the latent variables are continuous, as is common with variational autoencoders, one must discretize the density $q_{\bT}(\bz|\bx)$ into bins of volume $\delta \bz$, yielding a probability mass function $q_{\bT}(\bz|\bx) \delta \bz$. But this imposes a requirement on the initial bits: now $N_\mathrm{init} \geq -\log q_{\bT}(\bz|\bx) - \log \delta \bz$ increases as the discretization resolution $1 / \delta \bz$ increases.

\citet{townsend2018practical} remark that initial bits can be avoided by transmitting multiple datapoints in sequence, where every datapoint $\bx^i$ (except for the first one $\bx^1$) uses the bitstream built up thus far as initial bitstream. This \emph{amortizes} the initial cost when the number of datapoints transmitted is large, but the cost can be significant for few or moderate numbers of datapoints, as we will see in experiments in Section~\ref{sec:experiments}.

\section{Problem Scenario: Hierarchical Latent Variables}

\begin{algorithm}[t!]
   \caption{BB-ANS for lossless compression with hierarchical latent variables. The operations below show the procedure for encoding a dataset $\mathcal{D}$ onto a bitstream.}
   \label{alg:bbans}
\begin{algorithmic}
   \STATE {\bfseries Input:} data $\mathcal{D}$, depth $L$, $p_{\bT}(\bx, \bz_{1:L})$, $q_{\bT}(\bz_{1:L}|\bx)$
   \STATE {\bfseries Require:} ANS
   \STATE {\bfseries Initialize:} bitstream
   \REPEAT
   \STATE Take $\bx \in \mathcal{D}$
   \STATE decode $\bz_1$ with $q_{\bT}(\bz_1|\bx)$
   \FOR{$i=1$ {\bfseries to} $L-1$}
   \STATE decode $\bz_{i+1}$ with $q_{\bT}(\bz_{i+1}|\bz_{i})$
   \ENDFOR
   \STATE encode $\bx$ with $p_{\bT}(\bx|\bz_1)$
   \FOR{$i=1$ {\bfseries to} $L-1$}
   \STATE encode $\bz_{i}$ with $p_{\bT}(\bz_{i}|\bz_{i+1})$
   \ENDFOR
   \STATE encode $\bz_L$ with $p(\bz_L)$
   \UNTIL $\mathcal{D} = \varnothing$
   \STATE {\bfseries Send:} bitstream
\end{algorithmic}
\end{algorithm}

Initial bits issues also arise when the model has many latent variables. Models with multiple latent variables are more expressive in practice can more closely model $p_\mathrm{data}$, leading to better compression performance. But since $-\log q_{\bT}(\bz|\bx)$ generally grows with the dimension of $\bz$, adding more expressive power to the latent variable model via more latent variables will incur a larger initial bitstream for BB-ANS.



\begin{algorithm}[t!]
   \caption{\textbf{Bit-Swap (ours)} for lossless compression with hierarchical latent variables. The operations below show the procedure for encoding a dataset $\mathcal{D}$ onto a bitstream.}
   \label{alg:bitswap}
\begin{algorithmic}
   \STATE {\bfseries Input:} data $\mathcal{D}$, depth $L$, $p_{\bT}(\bx, \bz_{1:L})$, $q_{\bT}(\bz_{1:L}|\bx)$
   \STATE {\bfseries Require:} ANS
   \STATE {\bfseries Initialize:} bitstream
   \REPEAT
   \STATE Take $\bx \in \mathcal{D}$
   \STATE decode $\bz_1$ with $q_{\bT}(\bz_1|\bx)$
   \STATE encode $\bx$ with $p_{\bT}(\bx|\bz_1)$
   \FOR{$i=1$ {\bfseries to} $L-1$}
   \STATE decode $\bz_{i+1}$ with $q_{\bT}(\bz_{i+1}|\bz_{i})$
   \STATE encode $\bz_{i}$ with $p_{\bT}(\bz_{i}|\bz_{i+1})$
   \ENDFOR
   \STATE encode $\bz_L$ with $p(\bz_L)$
   \UNTIL $\mathcal{D} = \varnothing$
   \STATE {\bfseries Send:} bitstream
\end{algorithmic}
\end{algorithm}

\begin{figure}[H]
\begin{center}
\includegraphics[width=0.87\columnwidth]{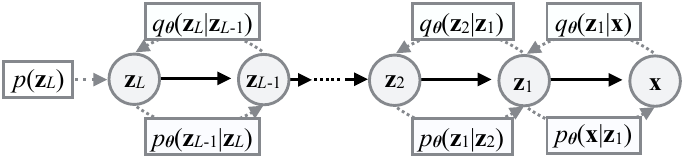}
\caption{The model class we are targeting: hierarchical latent variable models. Specifically, variational autoencoders whose sampling process obeys a Markov chain.}
\label{fig:multilatent}
\end{center}
\end{figure}

We specialize our discussion to the case of hierarchical latent variable models \cite{rezende2014stochastic}: variational autoencoders with multiple latent variables whose sampling process obeys a Markov chain of the form $\bz_L \rightarrow \bz_{L-1} \rightarrow \cdots \rightarrow \bz_1 \rightarrow \bx$, shown schematically in Figure~\ref{fig:multilatent}. (It is well known that such models are better density estimators than shallower models, 
and we will verify in experiments in Section~\ref{sec:experiments} that these models indeed can model $p_\mathrm{data}$ more closely than standard variational autoencoders. A discussion regarding other topologies can be found in Appendix \ref{appendix:future}.) Specifically, we consider a model whose marginal distributions are
\begin{align}
\begin{split} \label{eq:marginal}
   p_{\bT}(\bx) &= \int p_{\bT}(\bx|\bz_1)p_{\bT}(\bz_1) \text{d} \bz_1 \\
   p_{\bT}(\bz_1) &= \int p_{\bT}(\bz_1|\bz_2)p_{\bT}(\bz_2) \text{d} \bz_2 \\
   &\shortvdotswithin{=}
   p_{\bT}(\bz_{L-1}) &= \int p_{\bT}(\bz_{L-1}|\bz_{L})p(\bz_L) \text{d} \bz_L,
\end{split}
\end{align}
and whose marginal distribution over $\bx$ is
\begin{align}
   p_{\bT}(\bx) &= \int p_{\bT}(\bx|\bz_1)p_{\bT}(\bz_1|\bz_2) \dotsm p(\bz_L) \text{d} \bz_{1:L}. \label{eq:hiervaemarkov}
\end{align}
We define an inference model $q_{\bT}(\bz_i|\bz_{i-1})$ for every latent layer $\bz_i$, so that we can optimize a variational bound on the marginal likelihood $p_{\bT}(\bx)$. The resulting optimization objective (ELBO) is
\begin{equation}
    \loss(\bT) = \EX_{q_{\bT}(\cdot | \bx)}\lbrack{\log p_{\bT}(\bx, \bz_{1:L})-\log q_{\bT}(\bz_{1:L} | \bx)\rbrack}.
\end{equation}

\begin{figure*}[!htb]
\centering
\subfigure[\textbf{Bit-Swap (ours)}]{
\label{fig:bitswap}
\includegraphics[width=0.40\linewidth]{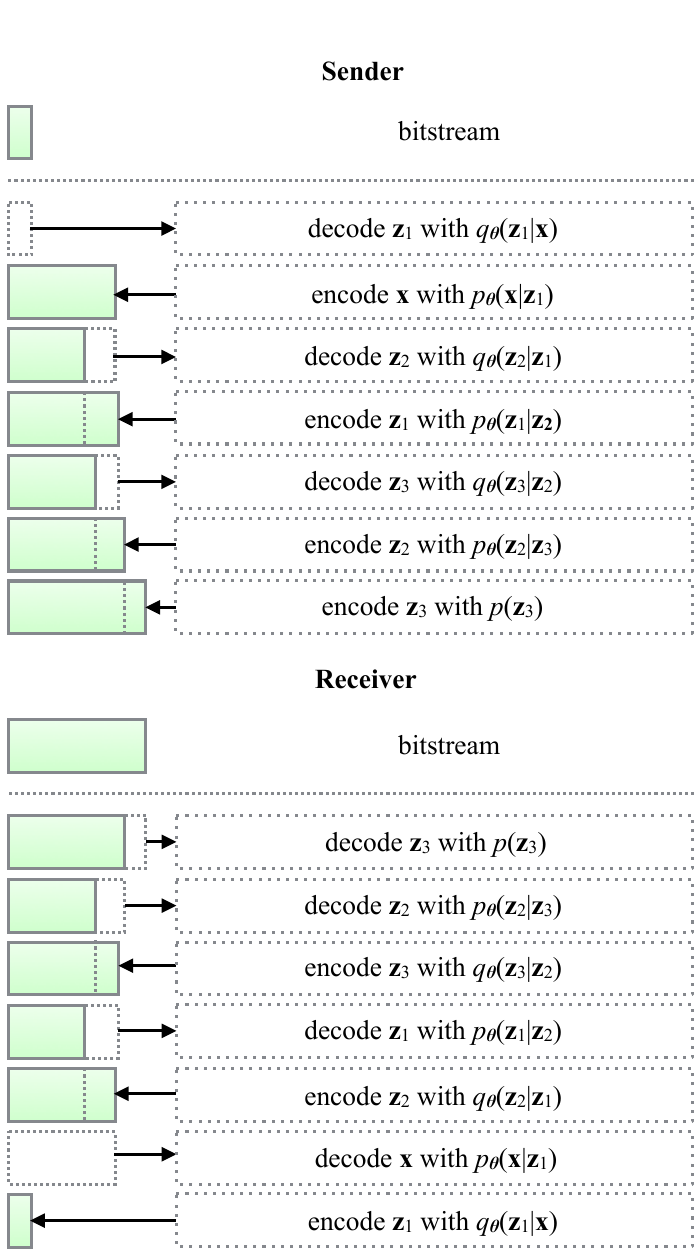}
}
\unskip\ \vrule\ 
\subfigure[BB-ANS]{
\label{fig:bbanshier}
\includegraphics[width=0.40\linewidth]{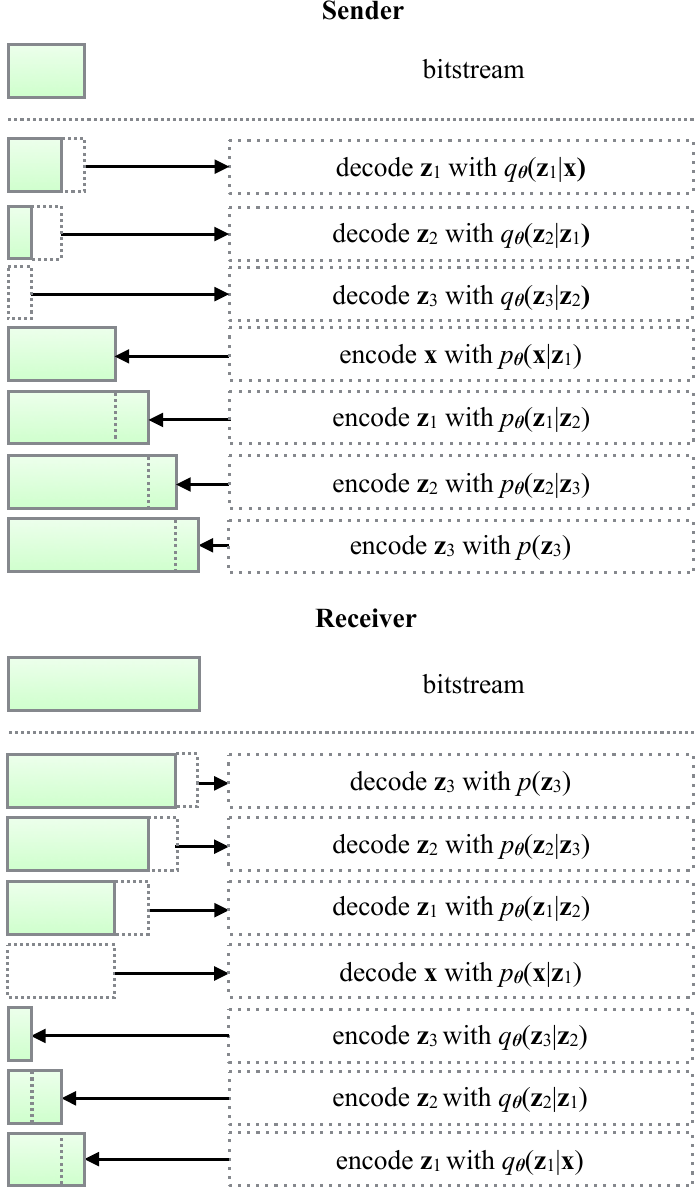}
}
\caption{\textbf{Bit-Swap (ours, left)} vs. BB-ANS (right) on a hierarchical latent variable model with three latent layers. Notice that BB-ANS needs a longer initial bitstream compared to Bit-Swap.}
\end{figure*}

Now, consider what happens when this model is used with BB-ANS for compression. Figure~\ref{fig:bbanshier} illustrates BB-ANS for such a model with three latent layers $\{\bz_1, \bz_2, \bz_3\}$; the algorithm for arbitrary latent depths $L$ of a hierarchical latent variable model is shown in Algorithm~\ref{alg:bbans}.

The first thing the sender must do is decode the latent variables $\bz_{1:L}$ from the initial bitstream of ANS. So, the number of bits present in the initial bitstream must be at least
\begin{align} \label{eq:overheadbbans}
N_\mathrm{init}^\mathrm{BBANS} \coloneqq -\log q_{\bT}(\bz_{1:L} | \bx) = \sum_{i=0}^{L-1} -\log q_{\bT}(\bz_{i+1} | \bz_{i})
\end{align}
where $\bz_0 = \bx$. Notice that $N_\mathrm{init}^\mathrm{BBANS}$ must grow with the depth $L$ of the latent variable model. With $L$ sufficiently large, the required initial bits could make BB-ANS impractical as a compression scheme with hierarchical latent variables.

\section{Bit-Swap}
\label{sec:bitswap}

To mitigate this issue, we propose Bit-Swap (Algorithm~\ref{alg:bitswap}), an improved compression scheme that makes bits-back coding efficiently compatible with the layered structure of hierarchical latent variable models.

In our proposed model~(Equations \ref{eq:marginal}-\ref{eq:hiervaemarkov}), the sampling process of both the generative model and the inference model obeys a Markov chain dependency between the stochastic variables. The data $\bx$ is generated conditioned on a latent variable $\bz_1$, as in a standard variational autoencoder. However, instead of using a fixed prior for $\bz_1$, we assume that $\bz_1$ is generated by a second latent variable $\bz_2$. Subsequently, instead of using a fixed prior for $\bz_2$, we assume that $\bz_2$ is generated by third latent variable $\bz_3$, and so on.

These nested dependencies enable us to recursively apply the bits-back argument as follows.
Suppose we aim to compress one datapoint $\bx$ in a lossless manner. 
With standard BB-ANS, the sender begins by decoding $\bz_{1:L}$, which incurs a large cost of initial bits. With Bit-Swap, we notice that we can apply the first two steps of the bits-back argument on the first latent variable: first decode $\bz_1$ and directly afterwards encode $\bx$. This adds bits to the bitstream, which means that further decoding operations for $\bz_{2:L}$ will need fewer initial bits to proceed. Now, we recursively apply the bits-back argument for the second latent variable $\bz_2$ in a similar fashion: first decode $\bz_2$ and afterwards encode $\bz_1$. Similar operations of encoding and decoding can be performed for the remaining latent variables $\bz_{3:L}$: right before decoding $\bz_{i+1}$, Bit-Swap always encodes $\bz_{i-1}$, and hence at least $-\log p_{\bT}(\bz_{i-1} | \bz_i)$ are available to decode $\bz_{i+1} \sim q_{\bT}(\bz_{i+1} | \bz_i)$ without an extra cost of initial bits. Therefore, the amount of initial bits that Bit-Swap needs is bounded by $\sum_{i=0}^{L-1} \max \left(0, \log \frac{p_{\bT}(\bz_{i-1}|\bz_i)}{q_{\bT}(\bz_{i+1} | \bz_i)} \right)$, where we used the convention $\bz_0=\bx$ and $p_{\bT}(\bz_{-1} | \bz_0) = 1$. We can guarantee that Bit-Swap requires no more initial bits than BB-ANS:
\begin{align}
N_\mathrm{init}^\mathrm{BitSwap} &\leq \sum_{i=0}^{L-1} \max \left(0, \log \frac{p_{\bT}(\bz_{i-1}|\bz_i)}{q_{\bT}(\bz_{i+1} | \bz_i)} \right) \\
 &\leq \sum_{i=0}^{L-1} -\log q_{\bT}(\bz_{i+1} | \bz_{i}) 
 = N_\mathrm{init}^\mathrm{BBANS}
\end{align}





See Figure~\ref{fig:bitswap} for an illustration of Bit-Swap on a model with three latent variables $\bz_1, \bz_2, \bz_3$.

\section{Experiments} \label{sec:experiments}

\begin{table*}[!ht]
\caption{\textbf{MNIST} model optimization results (columns 2 and 3) and test data compression results (columns 4 to 8) for various depths of the model (column 1). Column 2 shows the ELBO in bits/dim of the trained models evaluated on the test data. Column 3 denotes the number of parameters used (in millions). Using the trained models, we executed Bit-Swap and BB-ANS on the test data. We used $2^{10}$ bins to discretize the latent space (see Appendix~\ref{sec:discretization}). Column 5 denotes the scheme used; BB-ANS or Bit-Swap. Column 4 denotes the average net bitrate in bits/dim (see Section~\ref{sec:performance}), averaged over Bit-Swap and BB-ANS. Columns 6-8 show the cumulative moving average in bits/dim (CMA) (see Section~\ref{sec:performance}) at various timesteps (1, 50 and 100 respectively). The reported bitrates are the result of compression of 100 datapoints (timesteps), averaged over 100 experiments. We believe that the small discrepancy between the ELBO and the net bitrate comes from the noise resulting from discretization. Also, Bit-Swap reduces to BB-ANS for $L=1$.}
\label{tab:mnistresultsversus}
\begin{center}
\scriptsize
\begin{adjustbox}{width=0.79\linewidth}
\begin{tabular}{cccccccc}
\toprule
Depth ($L$) & ELBO $-\loss(\bT)$ & \# Parameters & Avg. Net Bitrate & Scheme & Initial ($n=1$) & CMA ($n=50$) & CMA ($n=100$) \\
\midrule
1& 1.35 & 2.84M & - & - & - & - & - \\ \hline \\
2& 1.28 & 2.75M &$1.28 \pm 0.34$ & BB-ANS & $6.59 \pm 0.30$ & $1.38 \pm 0.05$ & $1.33 \pm 0.03$\\
 &&&& \textbf{Bit-Swap} & $3.45 \pm 0.32$ & $1.32 \pm 0.05$ & $1.30 \pm 0.03$\\ \hline \\
4& 1.27 & 2.67M &$1.27 \pm 0.34$ & BB-ANS & $11.63 \pm 0.30$ & $1.47 \pm 0.05$ & $1.37 \pm 0.04$\\
 &&&& \textbf{Bit-Swap} & $3.40 \pm 0.31$ & $1.31 \pm 0.05$ & $1.29 \pm 0.04$\\ \hline \\
8& 1.27 & 2.60M &$1.27 \pm 0.33$ & BB-ANS & $21.93 \pm 0.34$ & $1.68 \pm 0.05$ & $1.48 \pm 0.03$\\
 &&&& \textbf{Bit-Swap} & $3.34 \pm 0.33$ & $1.31 \pm 0.05$ & $1.29 \pm 0.03$\\
\bottomrule
\end{tabular}
\end{adjustbox}
\end{center}
\end{table*}

\begin{table*}[!htb]
\caption{\textbf{CIFAR-10} model optimization (columns 2 and 3) and test data compression results (columns 4 to 8) for various depths of the model (column 1). Equal comments apply as Table \ref{tab:mnistresultsversus}.}
\label{tab:cifarresults}
\begin{center}
\scriptsize
\begin{adjustbox}{width=0.79\linewidth}
\begin{tabular}{cccccccc}
\toprule
Depth ($L$) & ELBO $-\loss(\bT)$ & \# Parameters & Avg. Net Bitrate & Scheme & Initial ($n=1$) & CMA ($n=50$) & CMA ($n=100$) \\
\midrule
1& 4.57 & 45.3M & - & - & - & - & - \\ \hline \\
2& 3.83 & 45.0M & $3.85 \pm 0.77$ & BB-ANS & $12.66 \pm 0.61$ & $4.03 \pm 0.11$ & $3.93 \pm 0.08$\\
 &&&& \textbf{Bit-Swap} & $6.76 \pm 0.63$ & $3.91 \pm 0.11$ & $3.87 \pm 0.08$\\ \hline \\
4& 3.81 & 44.9M &$3.82 \pm 0.83$ & BB-ANS & $22.30 \pm 0.83$ & $4.19 \pm 0.12$ & $4.00 \pm 0.09$\\
 &&&& \textbf{Bit-Swap} & $6.72 \pm 0.67$ & $3.89 \pm 0.12$ & $3.85 \pm 0.09$\\ \hline \\
8& 3.78 & 44.7M &$3.79 \pm 0.80$ & BB-ANS & $44.24 \pm 0.87$ & $4.60 \pm 0.12$ & $4.19 \pm 0.09$\\
 &&&& \textbf{Bit-Swap} & $6.53 \pm 0.74$ & $3.86 \pm 0.12$ & $3.82 \pm 0.09$\\
\bottomrule
\end{tabular}
\end{adjustbox}
\end{center}
\end{table*}

\begin{table*}[!htb]
\caption{\textbf{ImageNet ($32\times32$)} model optimization (columns 2 and 3) and test data compression results (columns 4 to 8) for various depths of the model (column 1). Equal comments apply as Table \ref{tab:mnistresultsversus}.}
\label{tab:imagenetresults}
\begin{center}
\scriptsize
\begin{adjustbox}{width=0.79\linewidth}
\begin{tabular}{cccccccc}
\toprule
Depth ($L$) & ELBO $-\loss(\bT)$ & \# Parameters & Avg. Net Bitrate & Scheme & Initial ($n=1$) & CMA ($n=50$) & CMA ($n=100$) \\
\midrule
1&4.94& 45.3M & - & - & - & - & - \\ \hline \\
2&4.53& 45.0M &$4.54 \pm 0.84$ & BB-ANS & $13.39 \pm 0.60$ & $4.71 \pm 0.13$ & $4.63 \pm 0.08$\\
 &&&& \textbf{Bit-Swap} & $7.45 \pm 0.62$ & $4.60 \pm 0.13$ & $4.57 \pm 0.08$\\ \hline \\
4&4.48& 44.9M &$4.48 \pm 0.85$ & BB-ANS & $22.72 \pm 0.79$ & $4.84 \pm 0.13$ & $4.66 \pm 0.08$\\
 &&&& \textbf{Bit-Swap} & $6.97 \pm 0.70$ & $4.53 \pm 0.13$ & $4.50 \pm 0.08$\\
\bottomrule
\end{tabular}
\end{adjustbox}
\end{center}
\end{table*}

\begin{figure*}[!htb]
\begin{center}
\includegraphics[width=0.86\textwidth]{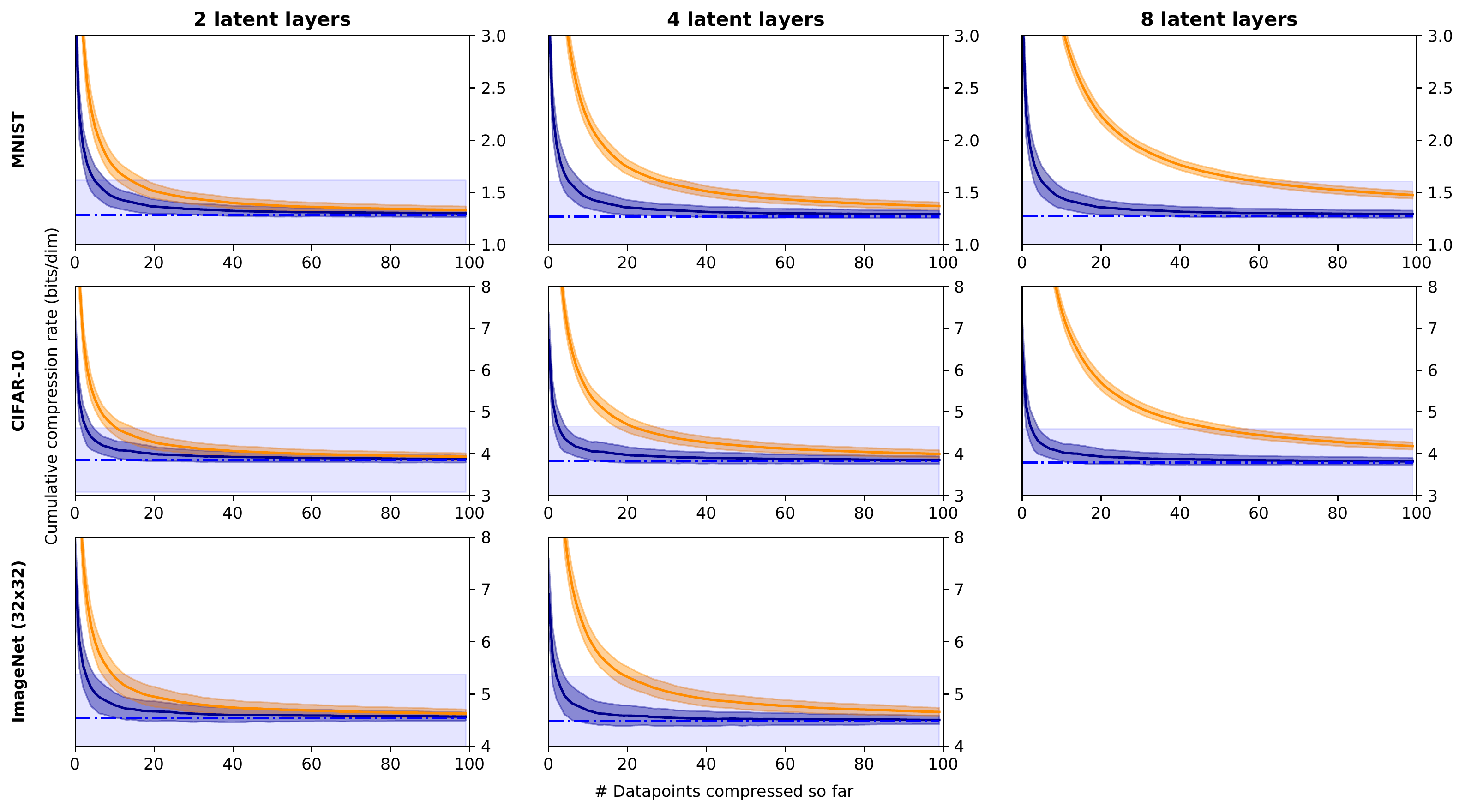}
\caption{Cumulative moving average of compression rate over time for Bit-Swap (blue) and BB-ANS (orange) for sequences of 100 datapoints, averaged over 100 experiments. The blue dotted line and region represent the average and standard deviation of the net bitrate across the entire test set, without the initial bits (see Section~\ref{sec:performance}). }
\label{fig:cma}
\end{center}
\end{figure*}

To compare Bit-Swap against BB-ANS, we use the following image datasets: MNIST, CIFAR-10 and ImageNet ($32\times32$). Note that the methods are not constrained to this specific type of data. As long as it is feasible to learn a hierarchical latent variable model with Markov chain structure $p_{\bT}(\bx)$ of the data under the given model assumptions, and the data is discrete, it is possible to execute the compression schemes Bit-Swap and BB-ANS on this data.

Referring back to the introduction, designing a good lossless compression algorithm is a matter of jointly solving two problems: \textbf{1)} approximating the true data distribution $p_{\text{data}}(\bx)$ as well as possible with a model $p_{\bT}(\bx)$, and \textbf{2)} developing a practical compression scheme that is compatible with this model and results in codelengths equal to $-\log p_{\bT}(\bx)$. We address the first point in Section \ref{performhier}. As for the second point, we achieve bitrates that are approximately equal to the $-\loss(\bT)$, the negative ELBO, which is an upper bound on $-\log p_{\bT}(\bx)$. We will address this in Section \ref{sec:performance}.

\subsection{Performance of Hierarchical Latent Variable Models} \label{performhier}
We begin our experiments by demonstrating how hierarchical latent variable models with Markov chain structure with different latent layer depths compare to a latent variable model with only one latent variable in terms of how well the models are able to approximate a true data distribution $p_{\text{data}}(\bx)$. A detailed discussion on the model architecture design can be found in Appendix~\ref{appendix:architecture}.

The results of training of the hierarchical latent variable models are shown in the left three columns of Table~\ref{tab:mnistresultsversus} (MNIST), \ref{tab:cifarresults} (CIFAR-10) and \ref{tab:imagenetresults} (ImageNet ($32\times32$)). One latent layer corresponds to \textit{one} latent variable $\bz_i$. The metric we used is bits per dimension (bits/dim) as evaluated by the negative ELBO $-L(\bT)$. Note from the resulting ELBO that, as we add more latent layers, the expressive power increases. A discussion on the utility of more latent layers can be found in Appendix \ref{appendix:posteriorcollapse}.

\subsection{Performance of Bit-Swap versus BB-ANS} \label{sec:performance}

We now show that Bit-Swap indeed reduces the initial bits required (as discussed in Section~\ref{sec:bitswap}) and outperforms BB-ANS on hierarchical latent variable models in terms of actual compression rates.
To compare the performance of Bit-Swap versus BB-ANS for different depths of the latent layers, we conducted 100 experiments for every model and dataset. In every experiment we compressed 100 datapoints in sequence and calculated the cumulative moving average (CMA) of the resulting lengths of the bitstream after each datapoint. Note that this includes the initial bits necessary for decoding latent layers.
In addition, we calculated the \textit{net} number of bits added to the bitstream after every datapoint, as explained in Section \ref{sec:bitsbackans}, and averaged them over all datapoints and experiments for one dataset and model. This can be interpreted as a lower bound of the CMA of a particular model and dataset. We discretized the continuous latent variables ${\bz}_{1:L}$ using $2^{10}$ discretization bins for all datasets and experiments, as explained in Appendix~\ref{sec:discretization}.

The CMA (with the corresponding average net bitrate) over 100 experiments for every model and dataset is shown in Figure~\ref{fig:cma}. Bit-Swap is depicted in blue and BB-ANS in orange. These graphs show two properties of Bit-Swap and BB-ANS: the difference between Bit-Swap and BB-ANS in the need for initial bits, and the fact that the CMA of Bit-Swap and BB-ANS both amortize towards the average net bitrate. The last five columns of Table~\ref{tab:mnistresultsversus} (MNIST), \ref{tab:cifarresults} (CIFAR-10) and \ref{tab:imagenetresults} (ImageNet ($32\times32$)) show the CMA (in bits/dim) after 1, 50 and 100 datapoints for the Bit-Swap versus BB-ANS and the average net bitrate (in bits/dim). 

The initial cost is amortized (see Section \ref{sec:bitsbackoverhead}) as the amount of datapoints compressed grows. Also, the CMA converges to the average net bitrate. The relatively high initial cost of both compression schemes comes from the fact that the initial cost increases with the number of discretization bins, discussed in Appendix \ref{sec:discretization}. Furthermore, discretizing the latent space adds noise to the distributions. When using BB-ANS, remember that this initial cost also grows linearly with the amount of latent layers $L$. Bit-Swap does not have this problem. This results in a CMA performance gap that grows with the amount of latent layers $L$. The efficiency of Bit-Swap compared to BB-ANS results in much faster amortization, which makes Bit-Swap a more practical algorithm.

Finally, we compared both Bit-Swap and BB-ANS against a number of benchmark lossless compression schemes. For MNIST, CIFAR-10 and Imagenet ($32\times32$) we report the bitrates, shown in Table~\ref{tab:benchmarks}, as a result of compressing 100 datapoints in sequence (averaged over 100 experiments) and used the best models reported in Table~\ref{tab:mnistresultsversus}, \ref{tab:cifarresults} and \ref{tab:imagenetresults} to do so. We also compressed 100 single images independently taken from the original unscaled ImageNet, cropped to multiples of 32 pixels on each side, shown in Table~\ref{tab:benchmarksfullimagenet}. First, we trained the same model as used for Imagenet ($32\times32$) on random $32\times32$ patches of the corresponding train set. Then we executed Bit-Swap and BB-ANS by compressing one $32\times32$ block at the time and averaging the bitrates of all the blocks in \textit{one} image. We used the same cropped images for the benchmark schemes. We did not include deep autoregressive models as benchmark, because they are too slow to be practical (see introduction). Bit-Swap clearly outperforms all other benchmark lossless compression schemes.

\begin{table}[tb]
\centering
\caption{Compression rates (in bits/dim) on MNIST, CIFAR-10, Imagenet ($32\times32$). The experimental set-up is explained in Section~\ref{sec:performance}.}
\label{tab:benchmarks}
\begin{adjustbox}{width=0.79\columnwidth}
\begin{tabular}{@{}llll@{}}
\toprule
\textbf{}                 & \textbf{MNIST} & \textbf{CIFAR-10} & \textbf{ImageNet} $(32 \times 32)$ \\ \midrule
Uncompressed              & 8.00           & 8.00              & 8.00                               \\ \midrule
GNU Gzip                  & 1.65           & 7.37              & 7.31                               \\
bzip2                     & 1.59           & 6.98              & 7.00                               \\
LZMA                      & 1.49           & 6.09              & 6.15                               \\
PNG                       & 2.80           & 5.87              & 6.39                               \\
WebP                      & 2.10           & 4.61              & 5.29                               \\
BB-ANS                      & 1.48           & 4.19              & 4.66                             \\
\textbf{Bit-Swap} & \textbf{1.29}  & \textbf{3.82}     & \textbf{4.50}                      \\ \bottomrule
\end{tabular}
\end{adjustbox}
\end{table}

\begin{table}[tb]
\centering
\caption{Compression rates (in bits/dim) on 100 images taken independently from unscaled and cropped ImageNet. The experimental set-up is explained in Section~\ref{sec:performance}.}
\label{tab:benchmarksfullimagenet}
\begin{adjustbox}{width=0.79\columnwidth}
\begin{tabular}{@{}ll@{}}
\toprule
\textbf{}                 & \textbf{ImageNet}\\ 
\textbf{}                 & (unscaled \& cropped) \\\midrule
Uncompressed                                             & 8.00              \\ \midrule
GNU Gzip \cite{gailly_adler_2018}                                               & 5.96              \\
bzip2 \cite{seward_2010}                                                   & 5.07              \\
LZMA \cite{pavlov_2019}                                                    & 5.09              \\
PNG                                                      & 4.71              \\
WebP                                                     & 3.66              \\
BB-ANS                                                   & 3.62              \\
\textbf{Bit-Swap}                                        & \textbf{3.51}     \\ \bottomrule
\end{tabular}
\end{adjustbox}
\end{table}
\section{Conclusion} \label{sec:conclusion}

Bit-Swap advances the line of work on practical compression using latent variable models, starting from the theoretical bits-back argument \citep{wallace1990classification,hinton1993keeping}, and continuing on to practical algorithms based on arithmetic coding \citep{frey1996free,frey1998bayesian} and asymmetric numeral systems \citep{townsend2018practical}.

Bit-Swap enables us to efficiently compress using hierarchical latent variable models with a Markov chain structure, as it is able to avoid a significant number of initial bits that BB-ANS requires to compress with the same models. The hierarchical latent variable models function as powerful density estimators, so combined with Bit-Swap, we obtain an efficient, low overhead lossless compression algorithm capable of effectively compressing complex, high-dimensional datasets. 

\clearpage
\section*{Acknowledgements}
We want to thank Diederik Kingma for helpful comments and fruitful discussions. This work was funded in part by ONR PECASE N000141612723, Huawei, Amazon AWS, and Google Cloud.






\bibliography{main}
\bibliographystyle{icml2019}

\clearpage
\appendix

\section{Asymmetric Numeral Systems (ANS)}
We will describe a version of Assymetric Numeral Systems that we have assumed access to throughout the paper and used in the experiments, namely the range variant (rANS). All the other versions and interpretations can be found in \cite{duda2009asymmetric}.

ANS encodes a (sequence of) data point(s) into a natural number $s \in \mathbb{N}$, which is called the \textit{state}. We will use unconventional notation, yet consistent with our work: $x$ to denote a single datapoint and $s$ to denote the state. The goal is to obtain a state $s$ whose length of the \textit{binary} representation grows with a rate that closely matches the \textit{entropy} of the data distribution involved. 

Suppose we wish to encode a datapoint $x$ that can take on two symbols $\{x_1, x_2\}$ that have equal probability. Starting with $s=1$. A valid scheme for this distribution is
\begin{equation}
\begin{split}
    x_1:& \ s \rightarrow 2s \\
    x_2:& \ s \rightarrow 2s + 1.
\end{split}
\end{equation}

This simply assigns 0 to $x_0$ and 1 to $x_1$ in binary. Therefore, it appends 1 or 0 to the right of the binary representation of the state $s$. Note that this scheme is fully decodable: if the current state is $s'$, we can read of the last encoded symbol by telling if the state $s'$ is even (last symbol was $x_1$) or odd (last symbol was $x_2$). Consequently, after figuring out which symbol $x$ was last encoded, the state $s$ before encoding that symbol is obtained by 
\begin{equation}
\begin{split}
    x_1 \ (s' \text{ even}):& \ s \rightarrow \frac{s'}{2} \\
    x_2 \ (s' \text{ odd}):& \ s \rightarrow \frac{s' - 1}{2}.
\end{split}
\end{equation}

Now, for the general case, suppose that the datapoint $x$ can take on a multitude of symbols $x \in \{x_1=1, x_2=2, ..., x_I=I\}$ with probability $\{p_1, p_2, ..., p_I\}$. In order to obtain a scheme analogous to the case with two symbols $\{x_1, x_2\}$, we have to assign every possible symbol $x_i$ to a specific \textit{subset of the natural numbers} $\mathcal{S}_i \subset \mathbb{N}$, that \textit{partitions} the natural numbers. Consequently, $\mathbb{N}$ is a disjoint union of the subsets $S_i$. Also, the elements in the subset $s \in S_i$ corresponding to $x_i$ must be chosen such that they occur in $\mathbb{N}$ with probability $p_i$.

This is accomplished by choosing a multiplier $M$, called the precision of ANS, that scales up the probabilities $\{p_1, p_2, ..., p_I\}$. The scaled up probability $p_i$ is denoted by $F[i]$ and the $F$'s are chosen such that $\sum_{i=1}^{I} F[i] = M$. We also choose subsets $\{K_1, K_2, ...\}$ that form intervals of length $M$ and partition the natural numbers. That means, the first $M$ numbers belong to $K_1$, the second $M$ numbers belong to $K_2$, and so on. Then, in every partition $K_n$, the first $M p_1$ numbers are assigned to symbol $x_1$ and form the subset $S_{n1}$, the second $M p_2$ numbers are assigned to symbol $x_2$ and form the subset $S_{n2}$, and so on. 

Now, we define $\mathcal{S}_i = \cup_{n=1}^{\infty} S_{ni}$. The resulting subsets $\mathcal{S}_i$ partition the natural numbers $\mathbb{N}$. Furthermore, the elements of $\mathcal{S}_i$ occur with probability approximately equal to $p_i$ in $\mathbb{N}$. 

Now, suppose we are given an initial state $s$. The scheme rANS can be interpreted as follows. \textbf{Encoding} a symbol $x_i$ is done by converting the state $s$ to a new state $s'$ that equals the $s^{\text{th}}$ occurrence in the set $\mathcal{S}_i$. This operation is made concrete in the following formula:
\begin{align*}
C(x,s) & = M \floor*{\frac{s}{F[x]}} + B[x] + s\Mod{F[x]}\\
& = s' \numberthis
\end{align*}
where $B[x] = \sum_{i=1}^{x-1} F[i]$, $R = s'\Mod{M}$ and $\floor*{\,}$ denotes the floor function.

Furthermore, suppose we are given a state $s'$ and we wish to know which number was last encoded (or in other words, we wish to \textbf{decode} from $s'$). Note that the union of the subsets $\mathcal{S}_i$ partitions the the natural numbers $\mathbb{N}$, so every number can be uniquely identified with one of the symbols $x_i$. Afterwards, if we know what the last encoded symbol $x_i$ was, we can figure out the state $s$ that \textit{preceded} that symbol by doing a look-up for $x_i$ in the set $\mathcal{S}_i$. The index of $x_i$ in $\mathcal{S}_i$ equals the state $s$ that preceded $s'$. This operation is made concrete in the following formula, which returns a symbol-state $(x,s)$ pair. 
\begin{align*}
    D(s') & = \left( \mathrm{argmax}\{B[x] < R\}, F[x] \floor{\frac{s'}{M}} + R - B[x] \right) \\
& = (x,s) \numberthis
\end{align*}

The new state $s'$ after encoding $x_i$ using this scheme is approximately equal to $\frac{s}{p_i}$. Consequently, encoding a sequence of symbols $x_1, x_2, ..., x_T$ onto the initial state $s$ results in a state $s_T$ approximately equal to $s_T \approx \frac{s}{p_1 p_2 \cdots p_T}$. Thus, the resulting codelength is
\begin{equation}
    \log s_T \approx \log s + \sum_t \log \frac{1}{p_T}
\end{equation}
If we divide by $T$, we obtain an average codelength which approaches the entropy of the data.

\section{The bits-back argument}
We will present a detailed explanation of the bits-back argument that is fitted to our case.
Again, suppose that a sender would like to communicate a sample $\bx$ to a receiver through a code that comprises the minimum amount of bits on average over $p_{\text{data}}$. Now suppose that both the sender and receiver have access to the distributions $q_{\bT}(\bz|\bx)$, $p(\bx|\bz)$ and $p(\bz)$, which parameters are optimized such that $p_{\bT}(\bx)$ $\big( = \int p_{\bT}(\bx|\bz)p(\bz) \text{d}\bz \big)$ approximates $p_{\text{data}}(\bx)$ well. Furthermore, both the sender and receiver have access to an entropy coding technique like ANS. 
A naive compression scheme for the \textbf{sender} is
\begin{enumerate}
    \item Sample $\bz \sim q_{\bT}(\bz|\bx)$
    \item Encode $\bx$ using $p_{\bT}(\bx|\bz)$
    \item Encode $\bz$ using $p(\bz)$.
\end{enumerate}
This would result in a bitrate equal to $N_{\text{total}} = -\log p_{\bT}(\bx|\bz) - \log p(\bz)$. The resulting bitstream gets sent over and the \textbf{receiver} would proceed in the following way:
\begin{enumerate}
    \item Decode $\bz$ using $p(\bz)$
    \item Decode $\bx$ using $p_{\bT}(\bx|\bz)$.
\end{enumerate}
Consequently, the sample $\bx$ is recovered in a lossless manner at the receiver's end. However, we can do better using the bits-back argument, which lets us achieve a bitrate of $\log q_{\bT}(\bz|\bx) - \log p_{\bT}(\bx|\bz) - \log p(\bz)$, which is equal to $N_{\text{total}} - \log q_{\bT}(\bz|\bx)$. To understand this, we have to clarify what \textit{decoding} means. If the model fits the true distributions perfectly, entropy coding can be understood as a (bijective) mapping between datapoints and uniformly random distributed bits. Therefore, assuming that the bits are uniformly random distributed, decoding information $\bx$ or $\bz$ from the bitstream can be interpreted as sampling $\bx$ or $\bz$. 

Now, assume the sender has access to an arbitrary bitstream $N_{\text{init}}$ that is already set in place. This can be in the form of previously compressed datapoints or other auxilary information. Looking at the naive compression scheme, we note that the step 'Sample $\bz \sim q_{\bT}(\bz|\bx)$' can be substituted by 'Decode $\bz$ using $q_{\bT}(\bz|\bx)$'. 
Consequently, the compression scheme at the \textbf{sender}'s end using the bits-back argument is
\begin{enumerate}
    \item Decode $\bz$ using $q_{\bT}(\bz|\bx)$
    \item Encode $\bx$ using $p_{\bT}(\bx|\bz)$
    \item Encode $\bz$ using $p(\bz)$.
\end{enumerate}
This results in a total bitrate equal to $N_{\text{total}} = N_{\text{init}} + \log q_{\bT}(\bz|\bx) - \log p_{\bT}(\bx|\bz) - \log p(\bz)$. The \textit{total} resulting bitstream gets sent over, and the \textbf{receiver} now proceeds as:
\begin{enumerate}
    \item Decode $\bz$ using $p(\bz)$.
    \item Decode $\bx$ using $p_{\bT}(\bx|\bz)$
    \item Encode $\bz$ using $q_{\bT}(\bz|\bx)$
\end{enumerate}
and, again, the sample $\bx$ is recovered in a lossless manner. But now, in the last step, the receiver has recovered the $N_{\text{init}}$ bits of auxiliary information that were set in place, thus gaining $N_{\text{init}}$ bits ``back". Ignoring the initial bits $N_{\text{init}}$, the \textit{net} number of bits regarding $\bx$ is
\begin{equation}
    N_{\text{total}} - N_{\text{init}} = \log q_{\bT}(\bz|\bx) - \log p_{\bT}(\bx|\bz) - \log p(\bz)
\end{equation}
which is on average equal to the negative ELBO $-\loss(\bT)$.

If the $N_{\text{init}}$ bits consist of relevant information, the receiver can then proceed by decompressing that information after gaining these "bits back". As an example, \citet{townsend2018practical} point out that it is possible to compress a sequence of datapoints $\{\bx_{1:N}\}$, where every datapoint $\bx_i$ (except for the first one $\bx_1$) uses the bitstream built up thus far as initial bitstream. Then, at the receiver's end, the datapoints $\{\bx_{1:N}\}$ get decompressed in reverse order. This way the receiver effectively gains the ``bits back" after finishing the three decompression steps of each datapoint $\bx_i$, such that decompression of the next datapoint $\bx_{i-1}$ can proceed. The only bits that can be irrelevant or redundant are the initial bits needed to compress the first datapoint $\bx_1$, though this information gets amortized when compressing multiple datapoints.

\section{AC and ANS: Queue vs. Stack} \label{appendix:queuestack}
\textit{Entropy coding} techniques make lossless compression possible given an arbitrary discrete distribution $p(\bx)$ over data. There exist several practical compression schemes, of which the most common flavors are
\begin{enumerate}
    \item Arithmetic Coding (AC) \cite{witten1987arithmetic}
    \item Asymmetric Numeral Systems (ANS). \cite{duda2009asymmetric}
\end{enumerate}
Both schemes operate by encoding the data into single number (in its binary representation equivalent to a sequence of bits or \textit{bitstream}) and decoding the other way around, where the single number serves as the message to be sent. By doing so, both schemes result in a message length with a small overhead of around 2 bits. That is, in case of compressing $\bx$, sending/receiving a corresponding codelength of approximately
\begin{equation}
    \EX{\lbrack -\log p(\bx) \rbrack} + 2 \ \  
    \text{bits}.
\end{equation}
However, AC and ANS differ in how they operate on the bitstream. In AC, the bitstream is treated as a \textit{queue} structure, where the first bits of the bitstream are the first to be decoded (FIFO). 

\begin{figure}[H]
\begin{center}
\includegraphics[width=0.87\columnwidth]{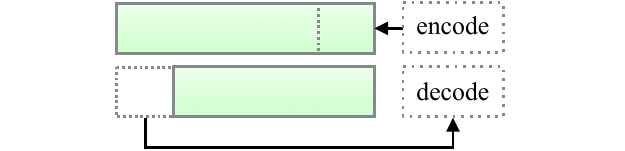}
\caption{Arithmetic Coding (AC) operates on a bitstream in a \textit{queue}-like manner. Symbols are decoded in the same order as they were encoded.}
\label{fig:acappendix}
\end{center}
\end{figure}
In ANS, the bitstream is treated as a \textit{stack} structure, where the last bits of the bitstream are the first to be decoded (LIFO).

\begin{figure}[H]
\begin{center}
\includegraphics[width=0.87\columnwidth]{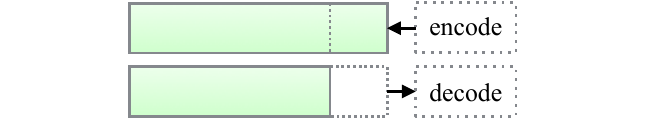}
\caption{Asymmetric Numeral Systems (ANS) operates on a bitstream in a \textit{stack}-like manner. Symbols are decoded in opposite order as they were encoded.}
\label{fig:ansappendix}
\end{center}
\end{figure}
AC and ANS both operate on one discrete symbol $x_d$ at the time. Therefore, the compression schemes are constrained to a distribution $p(\bx)$ that encompasses a product of discrete directed univariate distributions in order to operate. Furthermore, both the sender and receiver need to have access to the compression scheme \textit{and} the model in order to be able to communicate.
 
\citet{frey1996free} implement the bits-back theory on a single datapoint $\bx$ using AC. Then the \textit{net codelength} is equal to the ELBO, given the fact that we have access to an initial random bitstream. However, we must consider the length of the initial bitstream, which we call the \textit{initial bits}, from which we decode $\bz$ when calculating the \textit{actual codelength}, in which case the codelength degenerates to $\EX_{q_{\bT}(\cdot | \bx)}\lbrack{- \log p_{\bT}(\bx, \bz)\rbrack} \ \ \text{bits}$. So implementing bits-back on a single datapoint will not result in the advantage of getting ``bits back".

By communicating a sequence of datapoints $\mathcal{D}$, only the first datapoint $\bx^1$ needs to have an initial random bitstream set in place. Afterwards, a subsequent datapoint $\bx^i$ may just use the existing bitstream build up thus far to decode the corresponding latent variable $\bz^i$. This procedure was first described by \cite{frey1998bayesian}, and was called bits-back with feedback. We will use the shorter and more convenient term \textit{chaining}, which was introduced by \cite{townsend2018practical}.

Chaining is not directly possible with AC, because the existing bitstream is treated as a \textit{queue} structure. Whereas bits-back only works if the latent variable $\bz \sim p(\bz)$ is decoded earlier than the corresponding datapoint $\bx \sim p_{\bT}(\bx|\bz)$, demanding $\bz$ to be 'stacked' on top of $\bx$ when decoding. Frey solves this problem by reversing the order of bits of the encoded $(\bz, \bx)$ before adding it to the bitstream. This incurs a cost between 2 to 32 bits to the encoding procedure of each datapoint $\bx \in \mathcal{D}$, depending on implementation.

\section{Model Architecture}
\label{appendix:architecture}

\begin{table}[tb]
\caption{Hyperparameters of the model architecture of MNIST, CIFAR-10 and ImageNet ($32\times32$). The first three rows denote the dimensions of $\bx$, $\bz_i$ and the output of the used Residual blocks respectively. The fourth row marks the amount of latent layers $L$ used. The fifth and sixth row denote the amount of 'processing' Residual blocks $P$ and the 'ordinary' Residual blocks $B$ respectively, as explained in \ref{appendix:architecture}}.
\label{tab:architecture}
\begin{adjustbox}{width=\columnwidth}
\begin{tabular}{@{}llll@{}}
\toprule
                              & \textbf{MNIST} & \textbf{CIFAR-10} & \textbf{ImageNet ($32\times32$)} \\ \midrule
Dimension $\bx$ $(C, H, W)$         & $(1, 28, 28)$    & $(3, 32, 32)$       & $(3, 32, 32)$       \\
Dimension $\bz_i$ $(C, H, W)$       & $(1, 16, 16)$    & $(8, 16, 16)$       & $(8, 16, 16)$       \\
Dimension Residual $(C, H, W)$ & $(64, 16, 16)$   & $(256, 16, 16)$     & $(256, 16, 16)$     \\
Latent Layers ($L$) & $\{1, 2, 4, 8\}$   & $\{1, 2, 4, 8\}$     & $\{1, 2, 4\}$     \\
'Processing' Residual ($P$)       & 4               & 4                  & 4                 \\
'Ordinary' Residual ($B$)            & 8               & 8                 & 8                  \\ 
Dropout Rate ($p$)            & 0.2               & 0.3                 & 0.0                  \\ \bottomrule
\end{tabular}
\end{adjustbox}
\end{table}

\begin{figure}[tb]
\label{fig:archinference}
\begin{center}
\includegraphics[width=1.0\columnwidth]{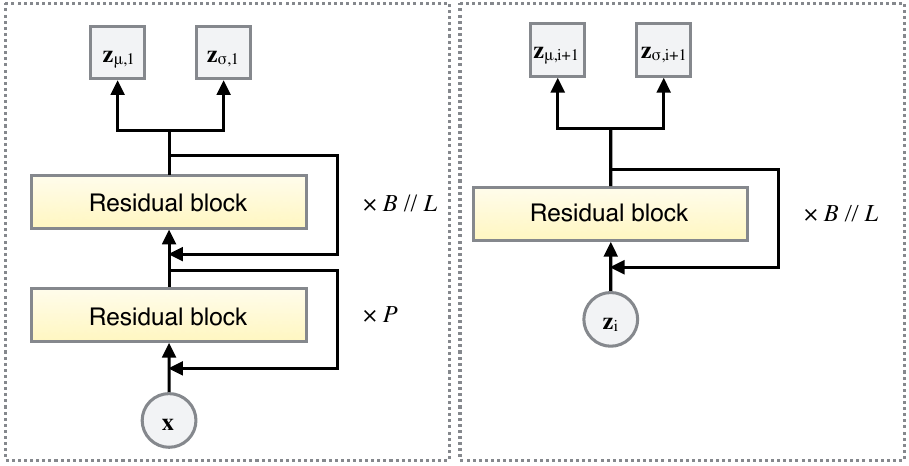}
\caption{A schematic representation of the networks corresponding to $q_{\bT}(\bz_1|\bx)$ (left) and $q_{\bT}(\bz_i|\bz_{i-1})$ (right) of the \textbf{inference} model. The arrows show the direction of the forward propagation.}
\label{fig:archinference}
\end{center}
\end{figure}

\begin{figure}[tb]
\begin{center}
\includegraphics[width=1.0\columnwidth]{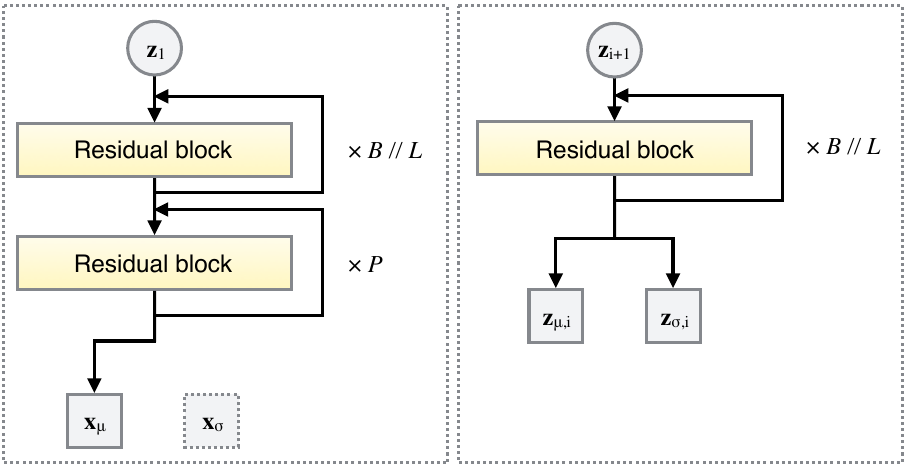}
\caption{A schematic representation of the networks corresponding to $p_{\bT}(\bx|\bz_1)$ (left) and $p_{\bT}(\bz_i|\bz_{i+l})$ (right) of the \textbf{generative} model. The arrows show the direction of the forward propagation.}
\label{fig:archgenerative}
\end{center}
\end{figure}

For all three datasets (MNIST, CIFAR-10 and ImageNet ($32\times32$)), we chose a Logistic distribution $(\mu=0,\sigma=1)$ for the prior $p(\bz_L)$ and conditional Logistic distributions for $q_{\bT}(\bz_i|\bz_{i-1})$, $q_{\bT}(\bz_1|\bx)$ and $p_{\bT}(\bz_i|\bz_{i+l})$. The distribution $p_{\bT}(\bx|\bz_1)$ is chosen to be a discretized Logistic distribution as defined in \cite{kingma2016improving}. We modeled the Logistic distributions by a neural network for every pair of parameters $(\mu,\sigma)$. A schematic representation of the different networks is shown in Figure~\ref{fig:archinference} and \ref{fig:archgenerative}. The $\sigma$ parameter of $p_{\bT}(\bx|\bz_1)$ is modeled unconditionally, and optimized directly. We chose Residual blocks \cite{he2015deep} as hidden layers. We also used Dropout~\cite{srivastava2014dropout} to prevent overfitting, Weight Normalization and Data-Dependent Initialization~\cite{salimans2016weight}, Polyak averaging \cite{polyak1992acceleration} of the model's parameters and the Adam optimizer \cite{kingma2015adam}.

To make a fair comparison between the different latent layer depths $L$ for one dataset, we used $B$ `ordinary' Residual blocks for the entire inference model and $B$ for the generative model, that is kept fixed for all latent layer depths $L$. The blocks are distributed over the $L$ networks that make up the inference model and $L$ networks that make up the generative model. In addition, we added $P$ `processing' Residual blocks at the beginning/end of the network corresponding to $q_{\bT}(\bz_1|\bx)$ and $p_{\bT}(\bx|\bz_1)$ respectively. Finally, we decreased the channel dimension of the output of all the Residual blocks in order to ensure that the parameter count stays constant (or regresses) as we add more latent layers. All the chosen hyperparameters are shown in Table~\ref{tab:architecture}. We refer to the code \url{https://github.com/fhkingma/bitswap} for further details on the implementation.

\section{Usefulness of Latent Layers \& Posterior Collapse}
\label{appendix:posteriorcollapse}

\begin{figure*}[!htb]
\begin{center}
\includegraphics[width=1.0\textwidth]{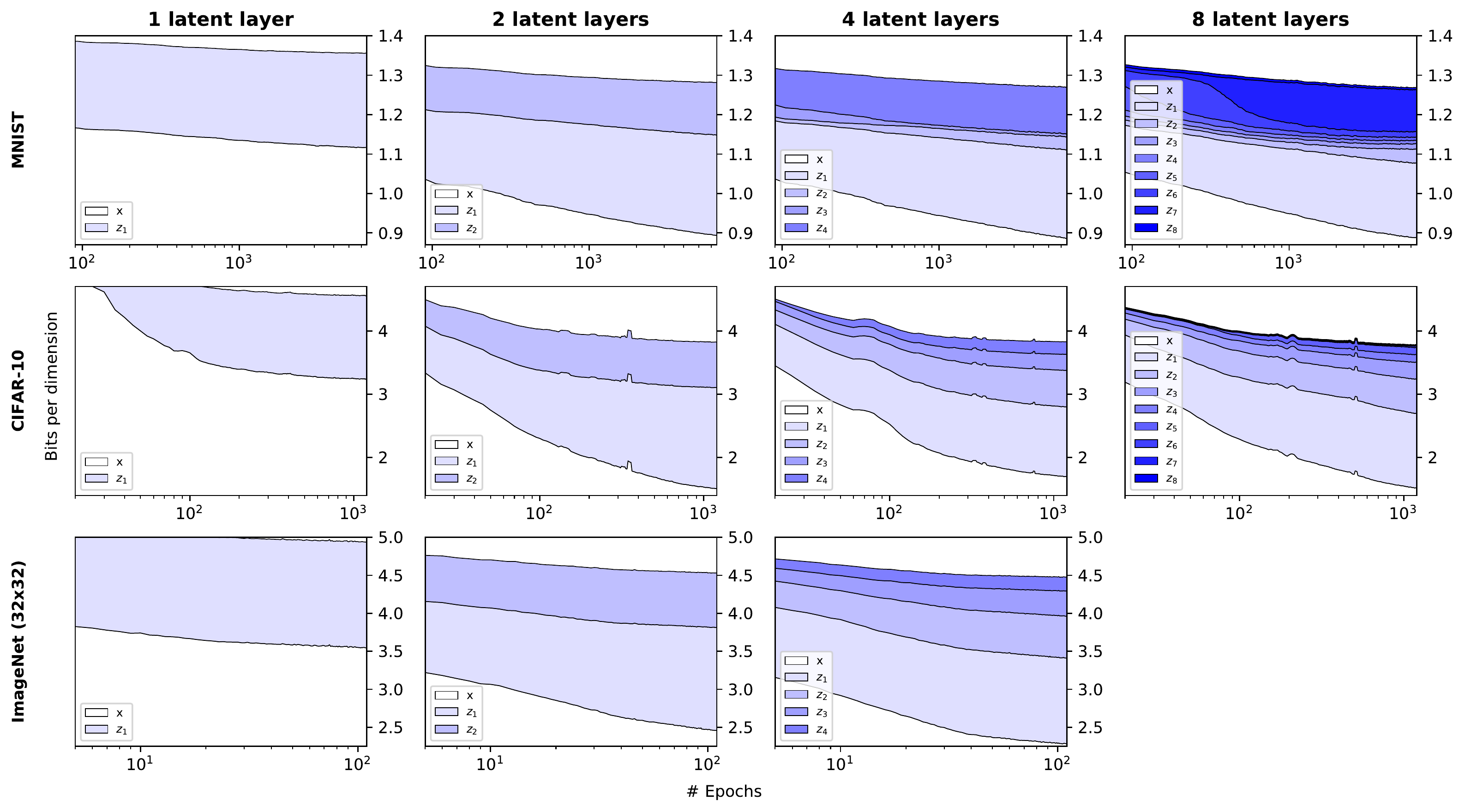}
\caption{Stack plots of the number of bits/dim required per stochastic layer to encode the test set over time. The bottom-most (white) area corresponds to the bottom-most (reconstruction of $\bx$) layer, the second area from the bottom denotes the first latent layer, the third area denotes the second latent layer, and so on.}
\label{fig:stackplot}
\end{center}
\end{figure*}

Posterior collapse is one of the drawbacks of using variational auto-encoders \cite{chen2016variational}. Especially when using deep hierarchies of latent variables, the higher latent layers can become redundant and therefore unused \cite{zhao2017deephierarchies}. We will counter this problem by using the free-bits technique as explained in \cite{chen2016variational} and \cite{kingma2016improving}. As a result of this technique, all latent layers across all models and datasets are used. To demonstrate this, we generated stack plots of the number of bits/dim required per stochastic layer to encode the test set over time shown in Figure~\ref{fig:stackplot}. The bottom-most (white) area corresponds to the bottom-most (reconstruction of $\bx$) layer, the second area from the bottom denotes the first latent layer, the third area denotes the second latent layer, and so on.

\section{Discretization of $\bz_{1:L}$} \label{sec:discretization}
In order to perform lossless compression with continuous latent distributions, we need to determine how to discretize the latent space $\bz_i$ for every corresponding distribution $p_{\bT}(\bz_i|\bz_{i+1})$, $q_{\bT}(\bz_{i}|\bz_{i-1})$ and $p(\bz_L)$. In \cite{townsend2018practical}, based on \cite{mackay2003information}, they show that if the bins $\delta_{\bz}$ of $\bz \sim p(\bz)$ match the bins $\delta_{\bz}$ of $\bz \sim q_{\bT}(\bz|\bx)$, continuous latents can be discretized up to arbitrary precision, without affecting the \textit{net} compression rate as a result of getting "bits back". We generalize this result to hierarchical latent variables by stating that the bins $\delta_{\bz_i}$ of the latent conditional generative distributions $(\bz_1, .., \bz_L) \sim p_{\bT}(\bz_1, .., \bz_L)$ have to match the bins $\delta_{\bz_i}$ of the inference distributions $(\bz_1, .., \bz_L) \sim q_{\bT}(\bz_1, .., \bz_L|\bx)$ in order to avoid affecting the compression rate. Nonetheless, the length of the initial bitstream needed to decode latent sample $\bz_1 \sim q_{\bT}(\bz_1|\bx)$ (or possibly samples $(\bz_1, .., \bz_L) \sim q_{\bT}(\bz_1, .., \bz_L|\bx)$) is still dependent on the corresponding bin size(s) $\delta_{\bz_i}$. Therefore, we cannot make the bin sizes $\delta_{\bz_i}$ too small without affecting the total codelength too much.

There are several discetization techniques we could use. One option is to simply discretize uniformly, which means dividing the space into bins of \textit{equal width}. However, given the constraint that the initial bitstream needed increases with larger precision, we have to make bin sizes reasonably large. Accordingly, uniform discretization of non-uniform distributions could lead to large discretization errors and this could lead to inefficient codelengths. 

An option is to follow the discretization technique used in \cite{townsend2018practical} by dividing the latent space into bins that have \textit{equal mass} under some distribution (as opposed to equal width). Ideally, the bins $\delta_{\bz_i}$ of $\bz_i \sim p_{\bT}(\bz_i|\bz_{i+1})$ match the bins $\delta_{\bz_i}$ of $\bz_i \sim q_{\bT}(\bz_i|\bz_{i-1})$ and the bins $\delta_{\bz_i}$ have equal mass under either $p_{\bT}(\bz_i|\bz_{i+1})$ or $q_{\bT}(\bz_i|\bz_{i-1})$. However, when using ANS with hierarchical latent variable models it is not possible to let the discretization of $\bz_i \sim q_{\bT}(\bz_i|\bz_{i-1})$ depend on bins based on $p_{\bT}(\bz_i|\bz_{i+1})$, because $\bz_{i+1}$ is not yet available for the \textit{sender} when decoding $\bz_i$. Conversely, discretization of $\bz_i \sim p_{\bT}(\bz_i|\bz_{i+1})$ cannot depend on bins based on $q_{\bT}(\bz_i|\bz_{i-1})$, since $\bz_{i-1}$ is not yet available for the \textit{receiver} decoding $\bz_i$. Note that the operations of the compression scheme at the sender end have to be the opposite of the operations at the receiver end and we need the \textit{same discretizations} for both ends. Under this conditions, it is not possible to use either $p_{\bT}(\bz_i|\bz_{i+1})$ or $p_{\bT}(\bz_i|\bz_{i-1})$ for the bin sizes $\delta_{\bz_i}$ and at the same time match the bins $\delta_{\bz_i}$ of $\bz_i \sim p_{\bT}(\bz_i|\bz_{i+1})$ with the bins $\delta_{\bz_i}$ of $\bz_i \sim q_{\bT}(\bz_i|\bz_{i-1})$.

So, we sampled a batch from the latent generative model $p_{\bT}(\bz_1, .., \bz_L)$ by ancestral sampling and a batch from the latent inference model (using the training dataset) right after learning. This gives us unbiased estimates of the statistics of the marginal distributions $p_{\bT}(\bz_1), .., p(\bz_L)$, defined in Equation \ref{eq:marginal}, which we can save as part of the model. Consequently, we used the marginal distributions to determine the bin sizes for discretization of $\bz_i \sim p_{\bT}(\bz_i|\bz_{i+1})$ and $\bz_i \sim q_{\bT}(\bz_i|\bz_{i-1})$. Note that we only have to perform this sampling operation once, hence this process does not affect the compression speed. 

However, we found that using uniform discretization for all latent layers $L$ except for the top one (corresponding to the prior) gives the best discretization and leads to the best compression results. Nevertheless, the top layer is discretized with bins that have equal mass under the prior, following \citet{townsend2018practical}.

\section{General Applicability of Bit-Swap}
\label{appendix:future}
Our work only concerns a very particular case of hierarchical latent variable models, namely hierarchical latent variable models in which the sampling process of both the generative- and inference model corresponding variational autoencoder obey Markov chains of the form 
\begin{equation*}
\bz_L \rightarrow \bz_{L-1} \rightarrow \cdots \rightarrow \bz_1 \rightarrow \bx
\end{equation*}
and
\begin{equation*}
    \bz_L \leftarrow \bz_{L-1} \leftarrow \cdots \leftarrow \bz_1 \leftarrow \bx
\end{equation*}
respectively. It might seem very restrictive to only be able to assume this topology of stochastic dependencies. However, the Bit-Swap idea can in fact be applied to any latent variable topology in which it is possible to apply the bits-back argument in a recursive manner. 

\begin{figure*}[h!]
\centering
\subfigure[Asymmetrical tree structure]{
\label{fig:tree}
\includegraphics[width=0.4\linewidth]{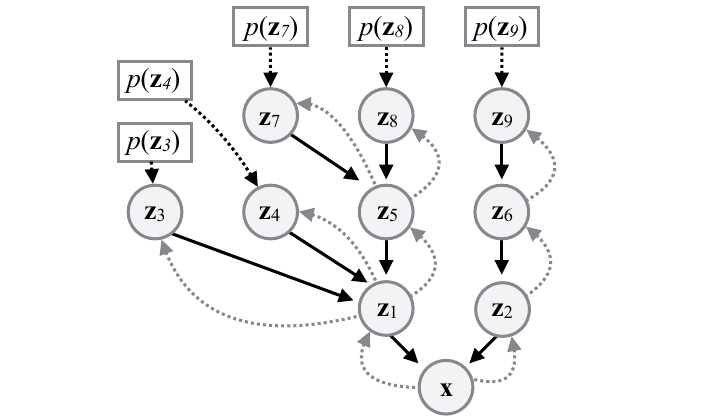}
}
\unskip\ \vrule\ 
\subfigure[Symmetrical tree structure including dependencies within a hierarchical layer]{
\label{fig:connectedtree}
\includegraphics[width=0.4\linewidth]{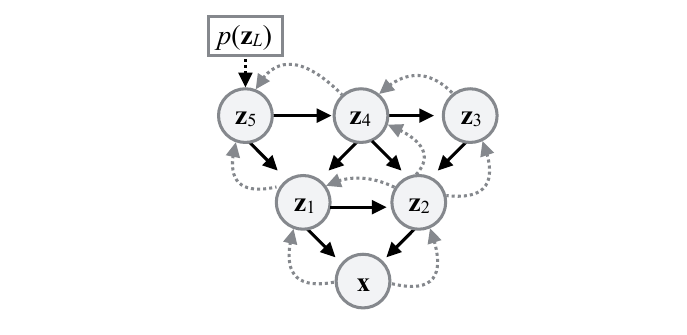}
}
\subfigure[Bit-Swap executed on \ref{fig:tree}]{
\label{fig:treebitswap}
\includegraphics[width=0.4\linewidth]{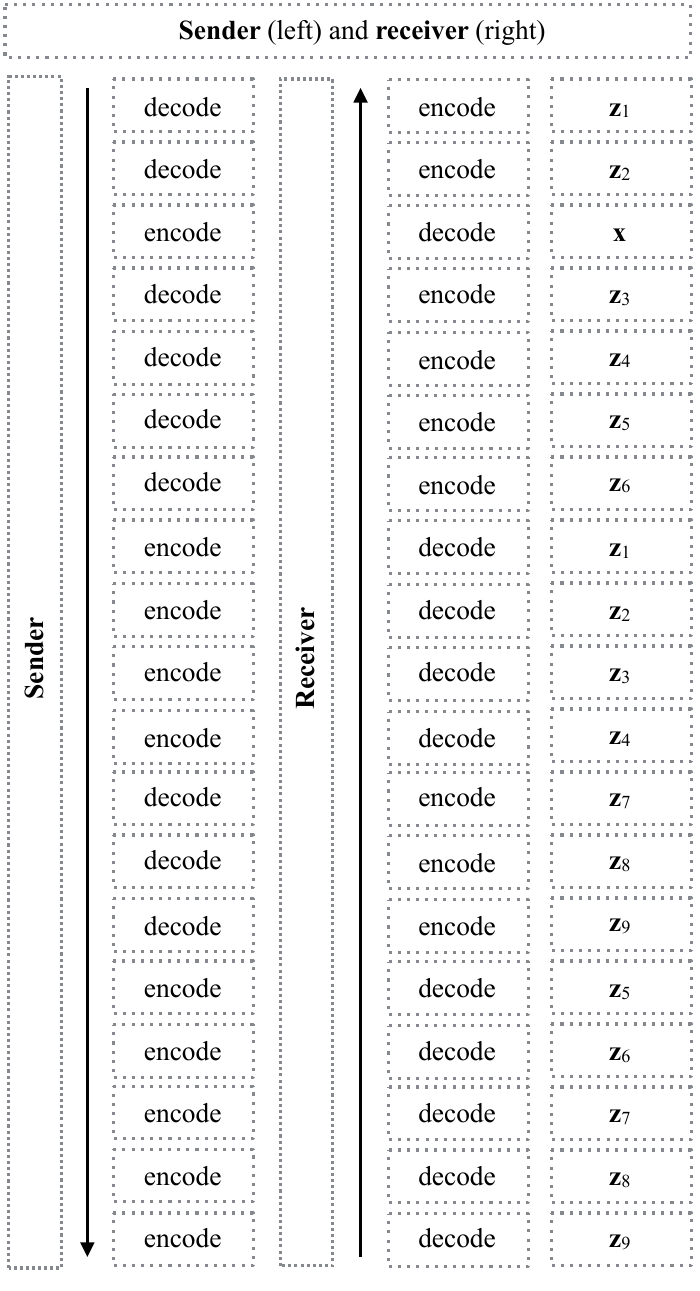}
}
\unskip\ \vrule\ 
\subfigure[Bit-Swap executed on \ref{fig:connectedtree}]{
\label{fig:connectedtreebitswap}
\includegraphics[width=0.4\linewidth]{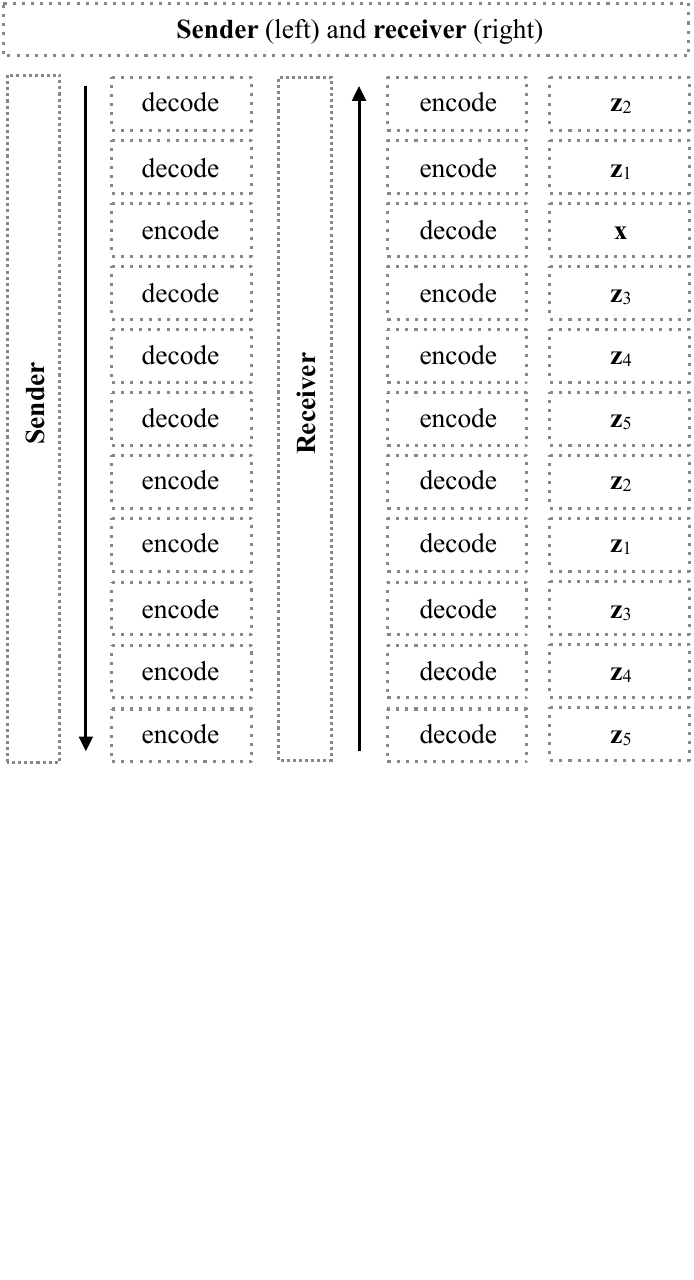}
}
\caption{The top left Figure shows an asymmetrical tree structure of stochastic dependencies and the top right Figure shows a symmetrical tree structure of stochastic dependencies where the variables of one hierarchical layer can also be connected. The black arrows indicate the direction of the generative model and the gray dotted arrows show the direction of the inference model. The black dotted arrows show where the prior(s) is/are defined on. In the bottom left and the bottom right we show the corresponding Bit-Swap compression schemes. In the right column of every Figure, we show the variables that are being operated on. On the left of every Figure we show the operations that must be executed by the sender and in the middle we show the operations executed by the receiver. The operations must be executed in the order that is dictated by the direction of the corresponding arrow. The sender always uses the inference model for decoding and the generative model for encoding. The receiver always uses the generative model for decoding and the inference model for encoding.}
\label{fig:topologies}
\end{figure*}

To show this we present two hypothetical latent variable topologies in Figure \ref{fig:topologies}. Figure \ref{fig:tree} shows an asymmetrical tree structure of stochastic dependencies and Figure \ref{fig:connectedtree} shows a symmetrical tree structure of stochastic dependencies where the variables of one hierarchical layer can also be connected. In Figure \ref{fig:treebitswap} and \ref{fig:connectedtreebitswap} we show the corresponding Bit-Swap compression schemes.

The more general applicability of Bit-Swap allows us to design complex stochastic dependencies and potentially stronger models. This might be an interesting direction for future work.

\end{document}